\documentclass[11pt]{article}

\usepackage{acl}

\usepackage{times}
\usepackage{latexsym}

\usepackage[T1]{fontenc}
\usepackage[utf8]{inputenc}
\usepackage{microtype}
\usepackage{inconsolata}
\usepackage{graphicx}
\usepackage{booktabs}
\usepackage{amsmath}
\usepackage{amssymb}
\usepackage{algorithm}
\usepackage{algorithmic}
\usepackage{xcolor}
\usepackage{multirow}
\usepackage{xspace}

\usepackage{tcolorbox}
\tcbuselibrary{skins,breakable,listings,theorems}

\newcommand{\method}{DARE\xspace}

\newcommand{\retriever}{Retriever\xspace}
\newcommand{\reasoner}{Reasoner\xspace}
\newcommand{\controller}{Controller\xspace}

\title{DARE: Dialectical Agentic Reasoning for \\Structured Knowledge Fact Checking}

\author{
  Yifei Li$^{1}$,
  Xiaohan Zheng$^{1}$,
  Wentao Qian$^{1}$, 
  Liansheng Zhuang$^{1}$\thanks{Corresponding author.} \\
  $^{1}$University of Science and Technology of China \\
  \texttt{yifeilee@mail.ustc.edu.cn} \\
  \texttt{lszhuang@ustc.edu.cn}
}

\begin{document}
\maketitle

\begin{abstract}
Structured knowledge fact checking aims to determine the truthfulness of natural language claims by reasoning over structured evidence. Recent program-generation approaches leverage large language models (LLMs) to generate executable graph reasoning programs, achieving strong performance on structured knowledge fact checking benchmarks. However, these methods remain limited by invalid relation generation, single-path reasoning that lacks self-correction, and biased evidence assessment that tends to overestimate supporting signals. We propose \textbf{D}ialectical \textbf{A}gentic \textbf{RE}asoning (\textbf{DARE}), a multi-agent framework that formulates structured knowledge fact checking as an iterative retrieve--reason--reflect process. DARE integrates relation-grounded evidence retrieval to constrain reasoning to valid structures, dialectical bidirectional verification to evaluate evidence from both supporting and refuting perspectives, and confidence-driven meta-reflection to dynamically determine whether additional evidence exploration is necessary. Extensive experiments demonstrate the effectiveness of DARE in structured knowledge fact checking, with an 8B backbone achieving 88.12\% accuracy and matching or surpassing GPT-4o-based program-generation baselines, which attests to the efficacy of dialectical agentic reasoning in eliciting the latent reasoning capabilities of LLMs.
\end{abstract}

\begin{figure}[t]
    \centering
    \includegraphics[width=0.85\linewidth]{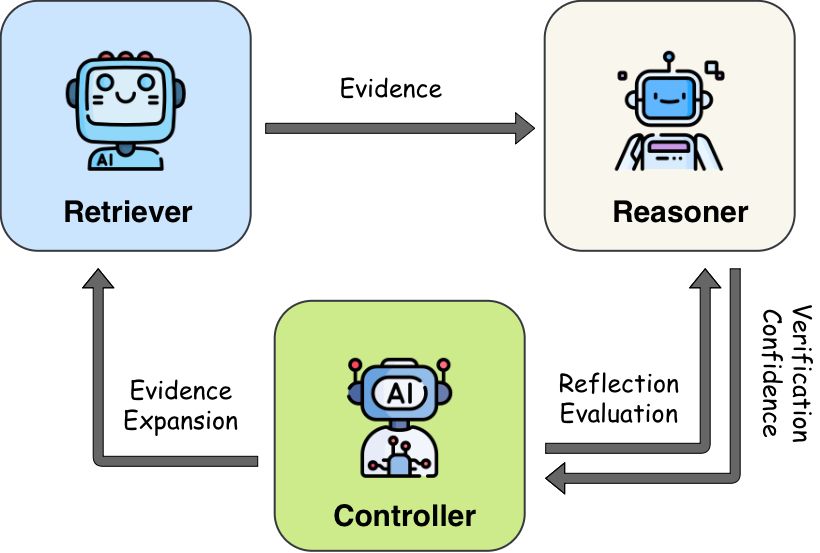}
    \caption{Interaction among the retriever, reasoner, and controller agents.}
    \label{fig:agent_interaction}
\end{figure}

\section{Introduction}

Structured knowledge fact checking aims to determine whether a natural language claim is supported or refuted by structured knowledge in knowledge graphs \citep{kim2023factkg,thorne2018fever}. Given a claim, the system must retrieve relevant evidence from the knowledge graph and produce a grounded binary decision. This task plays an important role in combating misinformation and has attracted increasing attention in the NLP community \citep{guo2022survey,nakov2021factcheck}.

Recent advances in large language models (LLMs) have shifted the paradigm from training specialized classifiers \citep{zhou2019gear} to program-generation approaches \citep{hao2025pgr,kim2023kggpt,pan2023programfc}, which prompt LLMs to generate executable reasoning programs composed of graph operations (e.g., \texttt{SEARCH}, \texttt{MATCH}, \texttt{VERIFY}) and execute them over the knowledge graph to obtain verification results.

Despite their success, we identify a fundamental limitation of existing program-generation methods: they produce ungrounded and uncorrected reasoning trajectories over knowledge graphs. In particular, reasoning failures arise from three tightly coupled issues. First, LLMs are unaware of relation vocabularies specific to knowledge graphs, leading to invalid relation generation and execution errors. Second, reasoning is performed in a single forward pass without mechanisms for iterative correction, making the process brittle under multi-hop or compositional claims. Third, verification is conducted from a single perspective, resulting in systematic overestimation of supporting evidence and poor confidence calibration.

To address this limitation, we propose \method (Dialectical Agentic REasoning), a unified dialectical reasoning framework that performs iterative, grounded verification over knowledge graphs. Instead of treating fact verification as one-shot program execution, \method formulates it as a closed-loop retrieve--reason--reflect process that couples evidence grounding with adversarial evaluation. The interaction among the three agents is illustrated in Figure~\ref{fig:agent_interaction}.

Specifically, \method first introduces relation-grounded retrieval, which constrains reasoning to valid KG structures by retrieving candidate relations from the graph and filtering them according to claim semantics, ensuring that all reasoning steps are grounded in actual KG relations. It then designs dialectical bidirectional reasoning, which evaluates evidence from both supporting and opposing perspectives to explicitly expose contradictions and reduce confirmation bias. Finally, it proposes confidence-driven meta-reflection, which monitors the agreement between the two perspectives and adaptively determines whether additional reasoning iterations are required, forming a unified loop that continuously refines both evidence and judgment.

\begin{itemize}
\item We identify ungrounded and uncorrected reasoning trajectories as a key limitation of program-generation methods for structured knowledge fact checking, and propose \method, a dialectical reasoning framework that addresses this issue through grounded retrieval, dialectical evaluation, and adaptive reflection.

\item We demonstrate that dialectical bidirectional reasoning improves evidence calibration, while confidence-driven meta-reflection enables iterative self-correction during inference, leading to more reliable multi-step reasoning.

\item On the FactKG benchmark, \method with a Qwen3-8B model achieves 88.12\% accuracy, matching or exceeding GPT-4o-based program-generation baselines, while using a substantially smaller backbone model.
\end{itemize}

\section{Related Work}

\paragraph{Fact Checking on Knowledge Graphs.}
Fact checking seeks to assess the truthfulness of claims based on structured or unstructured evidence \citep{thorne2018fever,guo2022survey}. Early benchmarks primarily rely on unstructured text \citep{thorne2018fever,aly2021feverous} or semi-structured tables \citep{chen2019tabfact}, leaving KG-based verification underrepresented until FactKG \citep{kim2023factkg}, which grounds claims in the DBpedia knowledge graph \citep{lehmann2015dbpedia} across multiple reasoning types.

Traditional approaches train specialized classifiers over graph structures \citep{zhou2019gear}. More recently, LLM-based methods have shifted toward few-shot and program-generation paradigms. KG-GPT \citep{kim2023kggpt} and StructGPT \citep{jiang2023structgpt} perform step-by-step reasoning via in-context learning, ProgramFC \citep{pan2023programfc} decomposes claims into verifiable sub-claims, and PGR \citep{hao2025pgr} defines graph reasoning primitives and prompts LLMs to compose executable programs. A parallel line of work explores LLM--KG integration for broader reasoning tasks: Think-on-Graph \citep{sun2024tog} iteratively explores KG paths through beam search, and RoG \citep{luo2024rog} generates faithful reasoning chains grounded in KG relations. Baek et al.\ \citep{baek2024kalm} augment LLMs with retrieved knowledge for verification but do not incorporate dialectical evaluation. Despite their differences, these methods largely rely on a single forward reasoning trajectory over the KG, without explicit mechanisms for iterative correction or counterfactual evaluation.

\paragraph{LLM Reasoning and Self-Reflection.}
LLM reasoning has progressed from chain-of-thought prompting \citep{wei2022cot,kojima2022large} to self-consistency \citep{wang2023selfconsistency}, structured search over multiple reasoning paths \citep{yao2024tot,besta2024got}, and reasoning-action integration \citep{yao2023react}. Self-reflection further enables models to critique and refine their outputs through verbal feedback \citep{shinn2023reflexion}, self-revision \citep{madaan2023selfrefine}, or tool-based critique mechanisms \citep{gou2024critic}. Chain-of-Verification \citep{dhuliawala2023cove} addresses hallucination by generating verification questions and checking consistency, though it operates over unstructured text rather than symbolic knowledge. Multi-agent systems extend this paradigm by enabling debate or role-based collaboration to improve factuality \citep{du2023improving,liang2023encouraging,sun2024llmqa}.

However, these approaches predominantly rely on either self-evaluation or consensus-seeking dynamics, both of which remain vulnerable to miscalibrated confidence and self-evaluation distortion \citep{kadavath2022calibration,tian2023justask}. Importantly, they lack structured counterfactual reasoning over external symbolic evidence such as knowledge graphs, which limits their effectiveness in verification tasks requiring grounded and adversarial evaluation.

\begin{figure*}[t]
    \centering
    \includegraphics[width=0.98\textwidth]{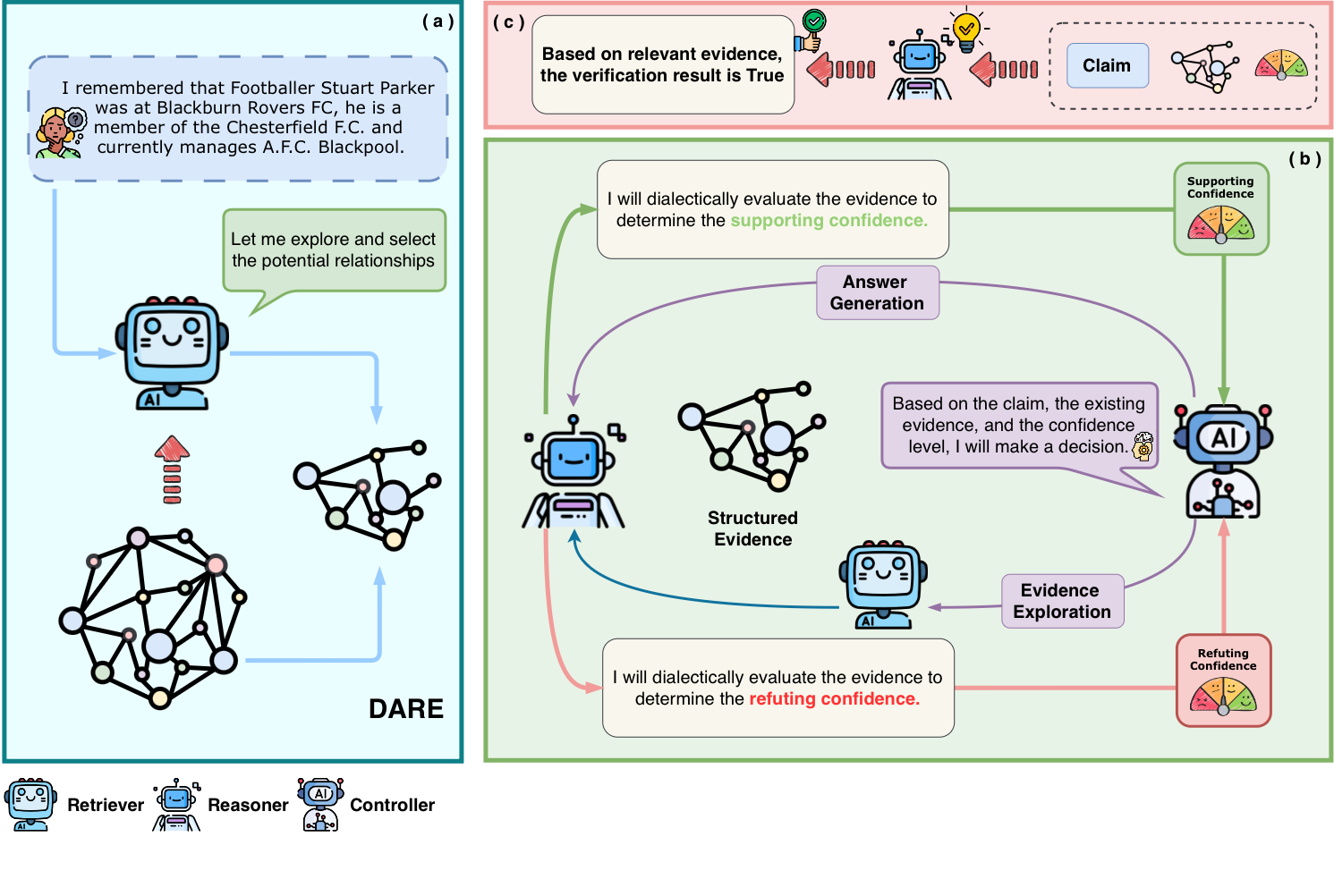}
    \caption{Overview of the DARE framework for structured knowledge fact checking. The retriever iteratively extracts relation-grounded evidence from the knowledge graph. The reasoner performs dialectical bidirectional reasoning over supporting and refuting perspectives to produce evidence-aware assessments. The controller applies confidence-driven meta-reflection to determine whether to terminate with a final prediction or continue iterative reasoning with expanded entities. }
    \label{fig:overview}
\end{figure*}

\section{Method}

\subsection{Task Formulation}
\label{sec:problem}

Given a knowledge graph $\mathcal{G} = (\mathcal{E}, \mathcal{R}, \mathcal{T})$, where $\mathcal{E}$ denotes the entity set, $\mathcal{R}$ the relation set, and $\mathcal{T} \subseteq \mathcal{E} \times \mathcal{R} \times \mathcal{E}$ the triple set, and a natural language claim $s$ mentioning a subset of entities $\mathcal{E}_s \subseteq \mathcal{E}$, the goal of KG fact verification is to predict a label:
\begin{equation}
  \label{eq:task}
  y = f(s, \mathcal{G}) \in \{\textsc{Sup}, \textsc{Ref}\},
\end{equation}
where $\textsc{Sup}$ and $\textsc{Ref}$ denote \emph{Supported} and \emph{Refuted}, respectively.

The task requires reasoning over the knowledge graph to determine whether the claim is supported by existing facts or contradicted by them. Figure~\ref{fig:overview} provides an overview of the proposed framework.

\subsection{Retrieval Module}
\label{sec:retriever}

The \retriever is responsible for gathering relevant evidence from the KG. To avoid the relation mapping failures inherent in program-generation approaches, the \retriever adopts a \emph{retrieve-then-filter} strategy that operates in two stages.

\paragraph{Adjacency Retrieval.}
For each entity $e \in \mathcal{E}_s$ mentioned in the claim, the \retriever retrieves all adjacent triples from $\mathcal{G}$ via direct neighborhood lookup:
\begin{equation}
  \label{eq:adjacency}
  \Gamma_e = \{(h, r, o) \in \mathcal{T} \mid h = e \lor o = e\},
\end{equation}
yielding the complete neighborhood subgraph $\Gamma_s = \bigcup_{e \in \mathcal{E}_s} \Gamma_e$.
This ensures that all retrieved relations exist in the KG, eliminating relation mapping errors. At iteration $t > 1$, the \retriever expands the entity set to include entities discovered in previously retrieved triples, enabling multi-hop evidence chains.

\paragraph{Semantic Filtering.}
The full adjacency set $\Gamma_s$ often contains relations irrelevant to verifying the claim. We employ an LLM-based evidence selector $G_E$ to semantically filter the most relevant triples based on the claim context:
\begin{equation}
  \label{eq:filter}
  E^{(t)} = G_E(\Gamma_s^{(t)}, s; \theta_E),
\end{equation}
where $\theta_E$ denotes the role-specific prompt that instructs the \retriever to select the top-$k$ most relevant triples from $\Gamma_s^{(t)}$ for verifying $s$.
Specifically, the LLM ranks adjacent triples by relevance and selects top-$k$ as $E^{(t)} \subseteq \Gamma_s^{(t)}$.
This retains only semantically relevant relations, reducing noise for subsequent reasoning while ensuring that all considered relations are valid KG entries.

\subsection{Dialectical Reasoning Module}
\label{sec:reasoner}

The \reasoner performs \emph{dialectical bidirectional reasoning} on the collected evidence to produce calibrated confidence assessments. Rather than evaluating evidence from a single perspective---which is susceptible to the self-evaluation distortion identified in Section~1---the \reasoner independently argues from both supporting and refuting viewpoints.

\paragraph{Dialectical Bidirectional Reasoning.}
Given the accumulated evidence $E_{1:t} = \bigcup_{i=1}^{t} E^{(i)}$ at iteration $t$, the \reasoner employs an LLM-based dialectical evaluator $D$ to assess the evidence from two opposing perspectives. Both branches take the same accumulated evidence $E_{1:t}$ as input but are guided by different prompts. First, the \reasoner generates a supporting argument $\alpha_t^{+}$ and produces a supporting confidence score:
\begin{equation}
  \label{eq:conf_pos}
  c_t^{+} = D^{+}(E_{1:t}, s; \theta_D^{+}),
\end{equation}
where $D^{+}$ denotes the supporting evaluation branch parameterized by prompt $\theta_D^{+}$, and $c_t^{+} \in [0, 1]$ quantifies the degree to which the evidence supports the claim. Independently, the \reasoner generates a refuting argument $\alpha_t^{-}$ and produces a refuting confidence score:
\begin{equation}
  \label{eq:conf_neg}
  c_t^{-} = D^{-}(E_{1:t}, s; \theta_D^{-}),
\end{equation}
where $D^{-}$ denotes the refuting evaluation branch parameterized by prompt $\theta_D^{-}$, and $c_t^{-} \in [0, 1]$ quantifies the degree to which the evidence refutes the claim.
The two prompts $\theta_D^{+}$ and $\theta_D^{-}$ share the same structure but carry opposing argumentation directives: $\theta_D^{+}$ instructs the model to argue why the evidence supports the claim, while $\theta_D^{-}$ instructs it to argue why the evidence refutes the claim.

By forcing the model to articulate both supporting and refuting arguments with equal rigor, dialectical reasoning counteracts self-evaluation distortion: even when evidence has a superficial supporting tendency, the model must confront potential contradictions, producing more balanced and reliable confidence estimates.

\paragraph{Cumulative Evidence Aggregation.}
At each iteration, the evidence and arguments from previous rounds are accumulated. The cumulative supporting and refuting confidences at iteration $t$ are computed as:
% \begin{equation}
%   \label{eq:aggregate}
%   C_t^{+} = \max_{1 \leq i \leq t} \, c_i^{+}, \quad C_t^{-} = \max_{1 \leq i \leq t} \, c_i^{-}, \quad
%   \alpha_t^{+*} = \alpha_{i^+}^{+}, \quad \alpha_t^{-*} = \alpha_{i^-}^{-},
% \end{equation}
\begin{equation}
\label{eq:agg_conf}
C_t^{+} = \max_{1 \leq i \leq t} c_i^{+}, \quad
C_t^{-} = \max_{1 \leq i \leq t} c_i^{-}.
\end{equation}

The corresponding best supporting and refuting arguments are selected from the iterations that achieve these maxima:
\begin{equation}
\label{eq:agg_arg}
\alpha_t^{+*} = \alpha_{i^+}^{+}, \quad
\alpha_t^{-*} = \alpha_{i^-}^{-},
\end{equation}
where $i^+ = \arg\max_{1 \leq i \leq t} c_i^{+}$ and $i^- = \arg\max_{1 \leq i \leq t} c_i^{-}$ denote the iteration indices yielding the highest supporting and refuting confidence scores, respectively.
Max-pooling captures the strongest supporting and refuting signals across all iterations, while $\alpha_t^{+*}$ and $\alpha_t^{-*}$ track the corresponding best arguments.
The best arguments are passed to the \controller for adjudication, ensuring that the final verdict is based on the strongest arguments rather than those from the most recent iteration alone.

\subsection{Meta-Controller}
\label{sec:controller}

The \controller implements a meta-reflection mechanism that determines whether the current evidence is sufficient for a final decision or whether further reasoning is required.

\paragraph{Confidence Gap.}
We use the confidence gap between the dialectical perspectives as the control signal:
\begin{equation}
  \Delta_t = |C_t^{+} - C_t^{-}|.
\end{equation}
A larger gap indicates more consistent evidence toward one verdict, while a smaller gap suggests ambiguity.

\paragraph{Meta-Reflective Decision.}
The \controller employs an LLM-based adjudication function that determines whether to stop reasoning or continue exploration:
% \begin{equation}
% \label{eq:meta_decision}
% \begin{aligned}
% \Phi(\alpha_t^{+*}, \alpha_t^{-*}, \Delta_t, \tau)
% =
% \begin{cases}
% \mathrm{Adj}(\alpha_t^{+*}, \alpha_t^{-*}; \theta_C), & \text{if } \Delta_t \geq \tau, \\
% \textsc{Continue}, & \text{otherwise}.
% \end{cases}
% \end{aligned}
% \end{equation}

\begin{equation}
\label{eq:meta_decision}
\Phi(\alpha_t^{+*}, \alpha_t^{-*}, \Delta_t, \tau)
=
\begin{cases}
\mathrm{Adj}(\cdot), & \Delta_t \geq \tau, \\
\textsc{Continue}, & \text{otherwise}.
\end{cases}
\end{equation}

Here, $\mathrm{Adj}(\cdot)$ denotes an LLM-based adjudicator that takes as input both the best supporting argument $\alpha_t^{+*}$ and the best refuting argument $\alpha_t^{-*}$, and produces a final decision by comparing their consistency with the accumulated evidence. The adjudicator synthesizes the two dialectical perspectives in a structured prompt and outputs the predicted label $\hat{y} \in \{\textsc{Sup}, \textsc{Ref}\}$.

If the maximum iteration $T$ is reached, the controller directly outputs the final verdict using the best available arguments:
\[
\hat{y} = \mathrm{Adj}(\alpha_T^{+*}, \alpha_T^{-*}; \theta_C).
\]

This design enables adaptive termination based on the discriminative strength of dialectical evidence, balancing efficiency and reasoning completeness.

\begin{algorithm}[t]
\caption{Dialectical Agentic REasoning}
\label{alg:admr}
\begin{algorithmic}[1]

\REQUIRE Claim $s$, KG $\mathcal{G} = (\mathcal{E}, \mathcal{R}, \mathcal{T})$, threshold $\tau$, maximum iterations $T$

\STATE Initialize: $C_0^{+} \leftarrow 0$, $C_0^{-} \leftarrow 0$, 
$E_{1:0} \leftarrow \emptyset$, $\alpha^{+*} \leftarrow \emptyset$, $\alpha^{-*} \leftarrow \emptyset$

\STATE Extract entity set $\mathcal{E}_s$ from $s$ via entity linking

\FOR{$t = 1, 2, \ldots, T$}

  \STATE \textbf{Retriever:} Retrieve adjacency set 
  $\Gamma_s^{(t)} \leftarrow \bigcup_{e \in \mathcal{E}_s} \{(h,r,o) \in \mathcal{T} \mid h=e \lor o=e\}$

  \STATE \textbf{Retriever:} Filter relevant evidence 
  $E^{(t)} \leftarrow G_E(\Gamma_s^{(t)}, s; \theta_E)$

  \STATE Update accumulated evidence:
  $E_{1:t} \leftarrow E_{1:t-1} \cup E^{(t)}$

  \STATE \textbf{Reasoner:} Generate supporting argument and score
  $c_t^{+}, \alpha_t^{+} \leftarrow D^{+}(E_{1:t}, s; \theta_D^{+})$

  \STATE \textbf{Reasoner:} Generate refuting argument and score
  $c_t^{-}, \alpha_t^{-} \leftarrow D^{-}(E_{1:t}, s; \theta_D^{-})$

  \STATE Update cumulative confidences:
  $C_t^{+} \leftarrow \max(C_{t-1}^{+}, c_t^{+}), \quad
   C_t^{-} \leftarrow \max(C_{t-1}^{-}, c_t^{-})$

  \STATE Update best arguments:
  $i^+ \leftarrow \arg\max_{1 \leq i \leq t} c_i^{+}, \quad
   \alpha^{+*} \leftarrow \alpha_{i^+}^{+}; \quad
   i^- \leftarrow \arg\max_{1 \leq i \leq t} c_i^{-}, \quad
   \alpha^{-*} \leftarrow \alpha_{i^-}^{-}$

  \STATE Compute confidence gap:
  $\Delta_t \leftarrow |C_t^{+} - C_t^{-}|$

  \IF{$\Delta_t \geq \tau$}
    \STATE \textbf{Controller:} 
    $\hat{y} \leftarrow \mathrm{Adj}(\alpha^{+*}, \alpha^{-*}; \theta_C)$
    \RETURN $\hat{y}$
  \ENDIF

  \STATE Expand entity set:
  $\mathcal{E}_s \leftarrow \mathcal{E}_s \cup \mathrm{Entities}(E^{(t)})$

\ENDFOR

\STATE \textbf{Controller:}
$\hat{y} \leftarrow \mathrm{Adj}(\alpha^{+*}, \alpha^{-*}; \theta_C)$

\RETURN $\hat{y}$

\end{algorithmic}
\end{algorithm}

\subsection{Overall Algorithm}
\label{sec:algorithm}

Algorithm~\ref{alg:admr} summarizes the complete \method framework. The procedure follows an iterative retrieve--reason--reflect paradigm, where evidence is progressively expanded and refined across iterations. Early iterations typically resolve simple cases (e.g., single-hop evidence suffices), whereas more complex claims require multiple rounds of retrieval and dialectical reasoning to accumulate discriminative evidence.

A key component of the framework is the dynamic expansion of the entity set based on newly retrieved triples, enabling multi-hop exploration over the knowledge graph. This mechanism allows the retriever to progressively traverse beyond initially mentioned entities, thereby supporting compositional reasoning over longer inference chains.

The iteration process is bounded by a maximum step $T$ to ensure computational efficiency. In practice, the controller typically converges within a small number of iterations (e.g., 2--3 steps), as the confidence gap criterion often triggers early termination when evidence becomes sufficiently discriminative.

\section{Experiments}

\subsection{Experimental Setup}

\paragraph{Dataset.}
We evaluate \method on the FactKG dataset \citep{kim2023factkg}, a benchmark designed for fact verification on knowledge graphs.
FactKG contains 108,764 claims based on factual information from the DBpedia knowledge graph \citep{lehmann2015dbpedia}.
Each instance consists of a claim, the corresponding DBpedia entities, a binary label (Supported or Refuted), and a reasoning type.
The test set comprises 9,041 claims spanning five reasoning types: one-hop (1,914), conjunction (3,069), existence (870), multi-hop (1,874), and negation (1,314), providing a comprehensive evaluation across varying claim complexities.

\paragraph{Baselines.}
Following the evaluation protocol of \citet{kim2023factkg} and \citet{hao2025pgr}, we categorize baselines into two groups based on evidence usage:

% \noindent\textbf{Without Evidence.} These methods rely solely on the claim text: BERT \citep{devlin2019bert}, BlueBERT \citep{liu2019roberta}, and Flan-T5 \citep{chung2024flant5} are transformer-based classifiers trained under the full-training strategy. ChatGPT uses 12-shot prompting without KG access.

\noindent\textbf{Without Evidence.}
These methods rely solely on the claim text:
BERT \citep{devlin2019bert} and BlueBERT \citep{bluebert2019}
are fine-tuned on the FactKG training set, whereas
Flan-T5 \citep{chung2024flant5} is evaluated in the zero-shot setting. ChatGPT uses 12-shot prompting.

\noindent\textbf{With Evidence.} These methods incorporate KG evidence: GEAR \citep{zhou2019gear} is a full-training model optimized for FactKG. KG-GPT \citep{kim2023kggpt} performs LLM-based KG reasoning under few-shot settings; we report results for the original gpt-3.5-turbo variant (KG-GPT), GPT-4o variant (KG-GPT$^{\dagger}$), GPT-4o-mini variant (KG-GPT$^{*}$), and DeepSeek-V3 variant (KG-GPT$_d$). ProgramFC \citep{pan2023programfc} generates verification programs. PGR \citep{hao2025pgr} is the strongest program-generation baseline; we report PGR (GPT-4o), PGR-mini (GPT-4o-mini), and PGR$_d$ (DeepSeek-V3).

\paragraph{Implementation Details.}
We implement \method using Qwen3-8B \citep{qwen2025qwen3} as the backbone LLM for all three agents. All agents share the same model but are instantiated with distinct role-specific prompts. The overall framework operates in a zero-shot setting without task-specific demonstrations.

The \retriever performs semantic filtering over retrieved KG triples using zero-shot prompting. The \reasoner is instructed with dialectical prompt that guides it to evaluate the evidence from both supporting and refuting perspectives and produce corresponding confidence scores in a unified output. The \controller uses an adjudication prompt that takes both dialectical arguments as input and outputs a final verdict. The confidence gap threshold is set to $\tau = 0.5$, and the maximum number of iterations is $T = 5$. The decoding hyperparameters are temperature = 0.1 and top-p = 0.9.

\subsection{Main Results}

\begin{table*}[t]
  \centering
  \footnotesize
  \setlength{\tabcolsep}{4pt}
  \begin{tabular}{llccccccc}
    \toprule
    \textbf{Method} & \textbf{Evidence} & \textbf{Strategy} & \textbf{1-Hop} & \textbf{Existence} & \textbf{Multi-Hop} & \textbf{Conjunction} & \textbf{Negation} & \textbf{Acc.} \\
    \midrule
    \multicolumn{9}{l}{\textit{Without Evidence}} \\
    BERT & w/o & full & 69.64 & 61.84 & 70.06 & 63.31 & 63.62 & 65.20 \\
    BlueBERT & w/o & full & 60.03 & 59.89 & 57.79 & 60.15 & 58.90 & 59.93 \\
    Flan-T5 & w/o & 0-shot & 62.17 & 55.29 & 60.67 & 69.66 & 55.02 & 62.70 \\
    ChatGPT & w/o & 12-shot & --- & --- & --- & --- & --- & 68.48 \\
    \midrule
    \multicolumn{9}{l}{\textit{With Evidence}} \\
    GEAR & w/ & full & 83.23 & 81.61 & 68.84 & 77.68 & 79.41 & 77.65 \\
    ProgFC & w/ & 12-shot & 88.11 & 64.42 & 63.13 & 87.16 & 57.61 & 74.62 \\
    KG-GPT & w/ & 12-shot & --- & --- & --- & --- & --- & 72.68 \\
    KG-GPT$^{*}$ & w/ & 12-shot & 80.56 & 63.19 & 55.84 & 69.93 & 59.19 & 67.04 \\
    KG-GPT$_d$ & w/ & 12-shot & 87.34 & 79.21 & 62.61 & 88.08 & 61.06 & 77.43 \\
    KG-GPT$^{\dagger}$ & w/ & 12-shot & 90.26 & 78.22 & 59.79 & 81.01 & 80.97 & 78.55 \\
    PGR-mini & w/ & 12-shot & 87.36 & 78.60 & 57.82 & 77.31 & 57.28 & 72.62 \\
    PGR$_d$ & w/ & 12-shot & 90.41 & 89.17 & 73.14 & 89.32 & 78.24 & 84.18 \\
    PGR & w/ & 12-shot & 93.30 & 93.10 & 75.96 & \textbf{89.95} & 81.16 & 86.82 \\
    \midrule
    \textbf{\method (Ours)} & w/ & \textbf{0-shot} & \textbf{94.44} & \textbf{95.23} & \textbf{77.28} & 87.58 & \textbf{88.89} & \textbf{88.12} \\
    \bottomrule
  \end{tabular}
  \caption{Accuracy (\%) on the FactKG dataset. Methods are categorized by evidence usage and training strategy. The best results are in \textbf{bold}. \method uses Qwen3-8B as the backbone model.}
  \label{tab:main}
\end{table*}

Table~\ref{tab:main} presents the experimental results. \method achieves the highest overall accuracy (88.12\%), outperforming PGR with GPT-4o (86.82\%) by 1.30\% despite using a model with substantially fewer parameters. This gap is primarily driven by two claim types where dialectical and iterative reasoning offer the most leverage.

On negation claims, \method reaches 88.89\%, a 7.73\% improvement over PGR's 81.16\%. Program-generation methods like PGR tend to map natural language negations directly into programmatic \texttt{NOT} operators, searching for missing edges rather than resolving the underlying logic. In contrast, the dialectical reasoner explicitly evaluates both supporting and refuting perspectives, forcing the model to confront potential contradictions instead of defaulting to a single-direction interpretation. A similar pattern emerges on existence claims (95.23\% vs.\ 93.10\%), where the retrieve-then-filter strategy ensures that all considered relations are grounded in the KG, eliminating the relation mapping errors that degrade program-generation baselines.

% On conjunction claims, \method (87.58\%) falls below PGR (89.95\%). Conjunction verification requires checking that every sub-claim holds simultaneously, a task well-suited to program composition, which can explicitly enumerate and verify each conjunct in a single structured execution. Our framework instead relies on iterative evidence accumulation and holistic confidence assessment, which may dilute the signal from individual conjuncts when they are evaluated together. This suggests that incorporating more structured sub-claim decomposition into the dialectical framework could further improve performance on compositional claim types.

On multi-hop claims, the most challenging type, \method achieves 77.28\%, surpassing PGR's 75.96\%. Multi-hop verification requires chaining multiple relational paths, and the iterative dialectical process enables the reasoner to accumulate and cross-validate intermediate evidence at each step before converging to a judgment. Across five reasoning types, \method outperforms PGR on four, achieving these gains under a 0-shot setting without any demonstrations—suggesting that structured agentic reasoning can effectively compensate for the absence of in-context examples.

\subsection{Ablation Study}

To understand the contribution of each agent, we conduct ablation experiments by removing the core mechanism of each agent respectively.

\begin{table}[t]
  \centering
  \footnotesize
  \setlength{\tabcolsep}{4pt}
  \begin{tabular}{lcc}
    \toprule
    \textbf{Variant} & \textbf{Acc.} & \textbf{$\Delta$} \\
    \midrule
    \method (Full) & \textbf{88.12} & --- \\
    \midrule
    w/o Reasoner's Dialectical Reasoning & 63.20 & $-$24.92 \\
    w/o Controller's Meta-Reflection & 70.11 & $-$18.01 \\
    w/o Retriever's Semantic Filtering & 72.21 & $-$15.91 \\
    \bottomrule
  \end{tabular}
  \caption{Ablation study on FactKG. Each variant removes the core mechanism of one agent. $\Delta$ shows the accuracy change relative to the full model.}
  \label{tab:ablation}
\end{table}

\paragraph{Reasoner's Dialectical Reasoning.}
Removing dialectical reasoning (i.e., using only a single-perspective evaluation instead of bidirectional scoring) causes the largest accuracy drop ($-$24.92\%).
This dramatic decline confirms that the dialectical structure is essential for calibrating confidence and combating self-evaluation distortion.
Without the opposing perspective, the model systematically overestimates supporting evidence, leading to a strong bias toward the \emph{Supported} verdict and severely undermining verification reliability.

\paragraph{Controller's Meta-Reflection.}
Removing meta-reflection (i.e., fixing the framework to a single iteration without dynamic termination) reduces accuracy by 18.01\%.
This confirms that iterative evidence gathering with confidence-driven termination is critical: complex claims such as multi-hop and negation require multiple rounds of evidence expansion and dialectical re-evaluation, which a single-pass pipeline cannot provide.

\paragraph{Retriever's Semantic Filtering.}
Removing semantic filtering (i.e., passing all adjacent triples directly to the \reasoner) reduces accuracy by 15.91\%.
While full adjacency retrieval prevents relation mapping failures, the resulting noise from irrelevant relations severely degrades the \reasoner's evaluation quality.
This highlights that precise evidence selection is as important as comprehensive retrieval---the \retriever must filter out irrelevant relations to enable effective dialectical reasoning.

\subsection{Analysis}

\begin{figure}[t]
    \centering
    \includegraphics[width=0.85\columnwidth]{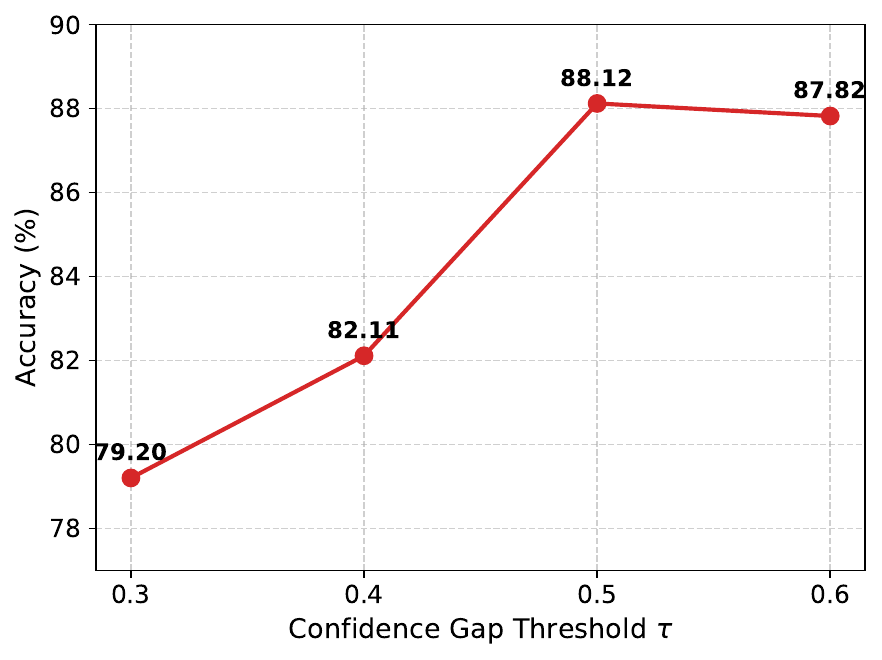}
    \caption{Impact of the confidence gap threshold ($\tau$) on the overall accuracy.}
    \label{fig:threshold}
\end{figure}

\paragraph{Effect of Confidence Gap Threshold.}
Figure~\ref{fig:threshold} shows the effect of varying the confidence gap threshold $\tau$.
When $\tau$ is too low (e.g., 0.3), the framework terminates prematurely, missing the opportunity to gather additional evidence for complex claims.
When $\tau$ is too high (e.g., 0.6), the framework requires excessively strong confidence before terminating, leading to unnecessary iterations that may introduce noise.
The optimal threshold $\tau = 0.5$ balances early termination for easy claims with sufficient exploration for complex ones.

\begin{figure}[t]
    \centering
    \includegraphics[width=0.85\columnwidth]{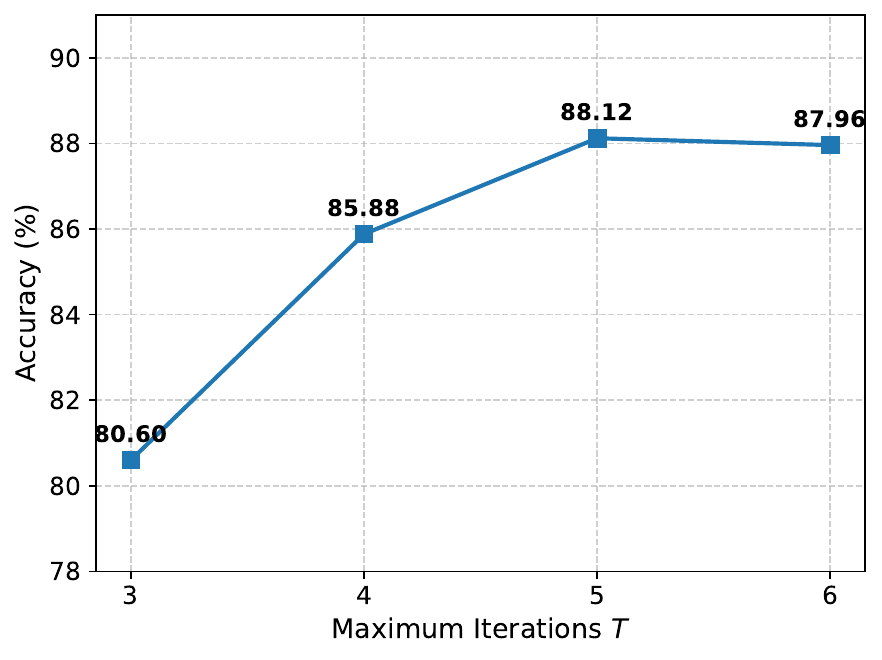}
    \caption{Impact of the maximum iteration limits ($T$) on the overall accuracy.}
    \label{fig:ablation_T}
\end{figure}

\paragraph{Effect of Maximum Iterations.}
Figure~\ref{fig:ablation_T} shows the effect of varying the maximum iteration limit $T$.
Accuracy improves substantially from $T=3$ (80.60\%) to $T=5$ (88.12\%), as additional iterations enable the framework to accumulate evidence for complex claims.
However, $T=6$ yields slightly lower accuracy (87.96\%), suggesting that excessive iterations introduce noise through unnecessary entity expansion.
We set $T=5$ as the default, which achieves the best performance while maintaining reasonable efficiency.

\paragraph{Case Study.}
Table~\ref{tab:case} illustrates how \method handles multi-hop claims that require evidence from multiple KG triples. The claim ``Aaron Boogaard was from Canada where one of the languages spoken is French'' involves a conjunction of two sub-claims: a birth-location fact and a language fact.

PGR (GPT-4o) generates a program that attempts to match both relations in a single execution, but fails because the strict programmatic syntax cannot map ``from'' to the exact valid KG paths, resulting in an execution failure and a \emph{False} prediction. In contrast, \method resolves this claim over two iterations. In the first iteration, the retriever extracts \texttt{(Aaron\_Boogaard, placeOfBirth, Canada)}, which supports the birth-location sub-claim but leaves the language sub-claim unverified. The reasoner assigns $c^{+} = 0.55$ and $c^{-} = 0.45$, yielding $\Delta = 0.10 < \tau$---insufficient to terminate. The controller then expands the entity set to include Canada, enabling the retriever in the second iteration to find \texttt{(Canada, language, French\_language)}. With both sub-claims now grounded, the reasoner updates its confidence to $c^{+} = 0.95$ and $c^{-} = 0.10$, producing $\Delta = 0.85 \ge \tau$. The controller adjudicates and outputs \emph{True}. This case demonstrates how iterative evidence expansion combined with confidence-driven termination allows \method to accumulate evidence across hops, avoiding the single-pass execution failures of program-generation baselines.

\begin{table}[t]
  \centering
  \footnotesize
  \setlength{\tabcolsep}{3pt}
  \renewcommand{\arraystretch}{1.1}
  \begin{tabular}{p{0.95\columnwidth}}
    \toprule
    \textbf{Claim:} Aaron Boogaard was from Canada where one of the languages spoken is French. \\
    \textbf{Label:} True \quad \textbf{Type:} Conjunction / Multi-claim \\
    \midrule
    \textbf{Baseline: PGR (GPT-4o)} \\
    \texttt{MATCH(Aaron\_Boogaard, placeOfBirth, ?x) AND MATCH(?x, language, French\_language)}. \\
    \textit{Error:} The strict programmatic syntax fails to map "from" to the exact valid KG paths in a single pass, leading to an execution failure. \\
    \textbf{Prediction:} \textbf{False} $\times$ \\
    \midrule
    \textbf{DARE (Ours)} \\
    \textbf{Iter.\ 1:} \\
    \textit{Retriever:} Extracts \texttt{(Aaron\_Boogaard, placeOfBirth, Canada)}. \\
    \textit{Reasoner:} $c^{+} = 0.55$ (Supports birth location, but missing language evidence), $c^{-} = 0.45$ (Refutes due to unverified language claim). \\
    \textit{Controller:} $\Delta = 0.10 < \tau$. Continues reasoning. \\
    \vspace{2pt}
    \textbf{Iter.\ 2:} \\
    \textit{Retriever:} Expands via `Canada` and extracts \texttt{(Canada, language, French\_language)}. \\
    \textit{Reasoner:} Cumulative evidence now satisfies both sub-claims. $c^{+} = 0.95$, $c^{-} = 0.10$. \\
    \textit{Controller:} $\Delta = 0.85 \ge \tau$. Terminates and adjudicates. \\
    \textbf{Prediction:} \textbf{True} $\checkmark$ \\
    \bottomrule
  \end{tabular}
  \caption{Case study on a multi-hop claim.}
  \label{tab:case}
\end{table}

\section{Conclusion}

We have presented \method, an agentic system for structured knowledge fact checking centered on the observation that LLMs suffer from self-evaluation distortion---a systematic bias toward overestimating supporting evidence when assessing claims from a single perspective. DARE mitigates this through dialectical bidirectional reasoning, which forces the model to argue from both supporting and refuting positions, producing calibrated confidence estimates rather than one-sided judgments. A confidence-gap-driven controller then determines whether evidence is sufficiently discriminative or whether the agents should continue exploring, enabling dynamic self-correction absent in single-pass methods. On FactKG, DARE with Qwen3-8B achieves state-of-the-art accuracy (88.12\%), surpassing PGR with GPT-4o while using a model substantially smaller. These findings indicate that addressing self-evaluation distortion through structured dialectical evaluation is a promising direction for improving LLM-based verification, and more broadly, that agentic designs which confront models with opposing evidence can yield more reliable reasoning over structured knowledge. We believe this principle extends beyond fact verification to other tasks requiring calibrated judgment over structured data, such as multi-hop reasoning and question answering. 

\section*{Limitations}

Despite the demonstrated effectiveness of DARE in structured knowledge fact checking, two limitations warrant further discussion. First, the \retriever's adjacency retrieval extracts all neighboring triples for each entity, which scales with the degree of the entity. For extremely dense knowledge graphs where hub entities possess thousands of adjacent relations, the initial retrieval step could introduce overhead, though semantic filtering subsequently reduces the set passed to downstream agents. Second, our empirical evaluation focuses on fact checking grounded in structured evidence. The results demonstrate the effectiveness of DARE in this setting, while its extension to a broader range of reasoning tasks holds substantial potential. Realizing this potential across diverse reasoning tasks represents an important direction for future work.

\bibliography{main}

@inproceedings{kim2023factkg,
    title = "{F}act{KG}: Fact Verification via Reasoning on Knowledge Graphs",
    author = "Kim, Jiho  and
      Park, Sungjin  and
      Kwon, Yeonsu  and
      Jo, Yohan  and
      Thorne, James  and
      Choi, Edward",
    editor = "Rogers, Anna  and
      Boyd-Graber, Jordan  and
      Okazaki, Naoaki",
    booktitle = "Proceedings of the 61st Annual Meeting of the Association for Computational Linguistics (Volume 1: Long Papers)",
    month = jul,
    year = "2023",
    address = "Toronto, Canada",
    publisher = "Association for Computational Linguistics",
    url = "https://aclanthology.org/2023.acl-long.895/",
    doi = "10.18653/v1/2023.acl-long.895",
    pages = "16190--16206"
}

@inproceedings{thorne2018fever,
    title = "{FEVER}: a Large-scale Dataset for Fact Extraction and {VER}ification",
    author = "Thorne, James  and
      Vlachos, Andreas  and
      Christodoulopoulos, Christos  and
      Mittal, Arpit",
    editor = "Walker, Marilyn  and
      Ji, Heng  and
      Stent, Amanda",
    booktitle = "Proceedings of the 2018 Conference of the North {A}merican Chapter of the Association for Computational Linguistics: Human Language Technologies, Volume 1 (Long Papers)",
    month = jun,
    year = "2018",
    address = "New Orleans, Louisiana",
    publisher = "Association for Computational Linguistics",
    url = "https://aclanthology.org/N18-1074/",
    doi = "10.18653/v1/N18-1074",
    pages = "809--819"
}

@inproceedings{aly2021feverous,
 author = {Aly, Rami and Guo, Zhijiang and Schlichtkrull, Michael and Thorne, James and Vlachos, Andreas and Christodoulopoulos, Christos and Cocarascu, Oana and Mittal, Arpit},
 booktitle = {Proceedings of the Neural Information Processing Systems Track on Datasets and Benchmarks},
 editor = {J. Vanschoren and S. Yeung},
 pages = {},
 title = {FEVEROUS: Fact Extraction and VERification Over Unstructured and Structured information},
 url = {https://datasets-benchmarks-proceedings.neurips.cc/paper_files/paper/2021/file/68d30a9594728bc39aa24be94b319d21-Paper-round1.pdf},
 volume = {1},
 year = {2021}
}

@inproceedings{chen2019tabfact,
  title={TabFact : A Large-scale Dataset for Table-based Fact Verification},
  author={Chen, Wenhu and Wang, Hongmin and Chen, Jianshu and Zhang, Yunkai and Wang, Hong and Li, Shiyang and Zhou, Xiyou and Wang, William Yang},
  booktitle = {International Conference on Learning Representations (ICLR)},
  address = {Addis Ababa, Ethiopia},
  month = {April},
  year = {2020}
}

@inproceedings{hao2025pgr,
    title = "Fact Verification on Knowledge Graph via Programmatic Graph Reasoning",
    author = "Hao, Yuanzhen  and
      Wu, Desheng",
    editor = "Christodoulopoulos, Christos  and
      Chakraborty, Tanmoy  and
      Rose, Carolyn  and
      Peng, Violet",
    booktitle = "Findings of the Association for Computational Linguistics: EMNLP 2025",
    month = nov,
    year = "2025",
    address = "Suzhou, China",
    publisher = "Association for Computational Linguistics",
    url = "https://aclanthology.org/2025.findings-emnlp.293/",
    doi = "10.18653/v1/2025.findings-emnlp.293",
    pages = "5480--5495",
    ISBN = "979-8-89176-335-7"
}

@inproceedings{kim2023kggpt,
    title = "{KG}-{GPT}: A General Framework for Reasoning on Knowledge Graphs Using Large Language Models",
    author = "Kim, Jiho  and
      Kwon, Yeonsu  and
      Jo, Yohan  and
      Choi, Edward",
    editor = "Bouamor, Houda  and
      Pino, Juan  and
      Bali, Kalika",
    booktitle = "Findings of the Association for Computational Linguistics: EMNLP 2023",
    month = dec,
    year = "2023",
    address = "Singapore",
    publisher = "Association for Computational Linguistics",
    url = "https://aclanthology.org/2023.findings-emnlp.631/",
    doi = "10.18653/v1/2023.findings-emnlp.631",
    pages = "9410--9421"
}

@inproceedings{zhou2019gear,
    title = "{GEAR}: Graph-based Evidence Aggregating and Reasoning for Fact Verification",
    author = "Zhou, Jie  and
      Han, Xu  and
      Yang, Cheng  and
      Liu, Zhiyuan  and
      Wang, Lifeng  and
      Li, Changcheng  and
      Sun, Maosong",
    editor = "Korhonen, Anna  and
      Traum, David  and
      M{\`a}rquez, Llu{\'i}s",
    booktitle = "Proceedings of the 57th Annual Meeting of the Association for Computational Linguistics",
    month = jul,
    year = "2019",
    address = "Florence, Italy",
    publisher = "Association for Computational Linguistics",
    url = "https://aclanthology.org/P19-1085/",
    doi = "10.18653/v1/P19-1085",
    pages = "892--901"
}

@inproceedings{pan2023programfc,
    title = "Fact-Checking Complex Claims with Program-Guided Reasoning",
    author = "Pan, Liangming  and
      Wu, Xiaobao  and
      Lu, Xinyuan  and
      Luu, Anh Tuan  and
      Wang, William Yang  and
      Kan, Min-Yen  and
      Nakov, Preslav",
    editor = "Rogers, Anna  and
      Boyd-Graber, Jordan  and
      Okazaki, Naoaki",
    booktitle = "Proceedings of the 61st Annual Meeting of the Association for Computational Linguistics (Volume 1: Long Papers)",
    month = jul,
    year = "2023",
    address = "Toronto, Canada",
    publisher = "Association for Computational Linguistics",
    url = "https://aclanthology.org/2023.acl-long.386/",
    doi = "10.18653/v1/2023.acl-long.386",
    pages = "6981--7004"
}

@inproceedings{jiang2023structgpt,
    title = "{S}truct{GPT}: A General Framework for Large Language Model to Reason over Structured Data",
    author = "Jiang, Jinhao  and
      Zhou, Kun  and
      Dong, Zican  and
      Ye, Keming  and
      Zhao, Xin  and
      Wen, Ji-Rong",
    editor = "Bouamor, Houda  and
      Pino, Juan  and
      Bali, Kalika",
    booktitle = "Proceedings of the 2023 Conference on Empirical Methods in Natural Language Processing",
    month = dec,
    year = "2023",
    address = "Singapore",
    publisher = "Association for Computational Linguistics",
    url = "https://aclanthology.org/2023.emnlp-main.574/",
    doi = "10.18653/v1/2023.emnlp-main.574",
    pages = "9237--9251"
}

@article{lehmann2015dbpedia,
  title={{DBpedia}: A Large-scale, Multilingual Knowledge Base Extracted from {Wikipedia}},
  author={Lehmann, Jens and Isele, Robert and Jakob, Max and Jentzsch, Anja and Kontokostas, Dimitris and Mendes, Pablo N and Hellmann, Sebastian and Morsey, Mohamed and Van Kleef, Patrick and Auer, S{\"o}ren and others},
  journal={Semantic web},
  volume={6},
  number={2},
  pages={167--195},
  year={2015},
  publisher={SAGE Publications Sage UK: London, England}
}

@article{kadavath2022calibration,
  title={Language models (mostly) know what they know},
  author={Kadavath, Saurav and Conerly, Tom and Askell, Amanda and Henighan, Tom and Drain, Dawn and Perez, Ethan and Schiefer, Nicholas and Hatfield-Dodds, Zac and DasSarma, Nova and Tran-Johnson, Eli and others},
  journal={arXiv preprint arXiv:2207.05221},
  year={2022}
}

@inproceedings{tian2023justask,
    title = "Just Ask for Calibration: Strategies for Eliciting Calibrated Confidence Scores from Language Models Fine-Tuned with Human Feedback",
    author = "Tian, Katherine  and
      Mitchell, Eric  and
      Zhou, Allan  and
      Sharma, Archit  and
      Rafailov, Rafael  and
      Yao, Huaxiu  and
      Finn, Chelsea  and
      Manning, Christopher",
    editor = "Bouamor, Houda  and
      Pino, Juan  and
      Bali, Kalika",
    booktitle = "Proceedings of the 2023 Conference on Empirical Methods in Natural Language Processing",
    month = dec,
    year = "2023",
    address = "Singapore",
    publisher = "Association for Computational Linguistics",
    url = "https://aclanthology.org/2023.emnlp-main.330/",
    doi = "10.18653/v1/2023.emnlp-main.330",
    pages = "5433--5442"
}

@article{qwen2025qwen3,
  title={Qwen3 technical report},
  author={Yang, An and Li, Anfeng and Yang, Baosong and Zhang, Beichen and Hui, Binyuan and Zheng, Bo and Yu, Bowen and Gao, Chang and Huang, Chengen and Lv, Chenxu and others},
  journal={arXiv preprint arXiv:2505.09388},
  year={2025}
}

@inproceedings{devlin2019bert,
    title = "{BERT}: Pre-training of Deep Bidirectional Transformers for Language Understanding",
    author = "Devlin, Jacob  and
      Chang, Ming-Wei  and
      Lee, Kenton  and
      Toutanova, Kristina",
    editor = "Burstein, Jill  and
      Doran, Christy  and
      Solorio, Thamar",
    booktitle = "Proceedings of the 2019 Conference of the North {A}merican Chapter of the Association for Computational Linguistics: Human Language Technologies, Volume 1 (Long and Short Papers)",
    month = jun,
    year = "2019",
    address = "Minneapolis, Minnesota",
    publisher = "Association for Computational Linguistics",
    url = "https://aclanthology.org/N19-1423/",
    doi = "10.18653/v1/N19-1423",
    pages = "4171--4186"
}

@inproceedings{bluebert2019,
    title = "Transfer Learning in Biomedical Natural Language Processing: An Evaluation of {BERT} and {ELM}o on Ten Benchmarking Datasets",
    author = "Peng, Yifan  and
      Yan, Shankai  and
      Lu, Zhiyong",
    editor = "Demner-Fushman, Dina  and
      Cohen, Kevin Bretonnel  and
      Ananiadou, Sophia  and
      Tsujii, Junichi",
    booktitle = "Proceedings of the 18th BioNLP Workshop and Shared Task",
    month = aug,
    year = "2019",
    address = "Florence, Italy",
    publisher = "Association for Computational Linguistics",
    url = "https://aclanthology.org/W19-5006/",
    doi = "10.18653/v1/W19-5006",
    pages = "58--65"
}

@article{chung2024flant5,
  author  = {Hyung Won Chung and Le Hou and Shayne Longpre and Barret Zoph and Yi Tay and William Fedus and Yunxuan Li and Xuezhi Wang and Mostafa Dehghani and Siddhartha Brahma and Albert Webson and Shixiang Shane Gu and Zhuyun Dai and Mirac Suzgun and Xinyun Chen and Aakanksha Chowdhery and Alex Castro-Ros and Marie Pellat and Kevin Robinson and Dasha Valter and Sharan Narang and Gaurav Mishra and Adams Yu and Vincent Zhao and Yanping Huang and Andrew Dai and Hongkun Yu and Slav Petrov and Ed H. Chi and Jeff Dean and Jacob Devlin and Adam Roberts and Denny Zhou and Quoc V. Le and Jason Wei},
  title   = {Scaling Instruction-Finetuned Language Models},
  journal = {Journal of Machine Learning Research},
  year    = {2024},
  volume  = {25},
  number  = {70},
  pages   = {1--53},
  url     = {http://jmlr.org/papers/v25/23-0870.html}
}

@inproceedings{wei2022cot,
 author = {Wei, Jason and Wang, Xuezhi and Schuurmans, Dale and Bosma, Maarten and Ichter, Brian and Xia, Fei and Chi, Ed and Le, Quoc V and Zhou, Denny},
 booktitle = {Advances in Neural Information Processing Systems},
 doi = {10.52202/068431-1800},
 editor = {S. Koyejo and S. Mohamed and A. Agarwal and D. Belgrave and K. Cho and A. Oh},
 pages = {24824--24837},
 publisher = {Curran Associates, Inc.},
 title = {Chain-of-Thought Prompting Elicits Reasoning in Large Language Models},
 url = {https://proceedings.neurips.cc/paper_files/paper/2022/file/9d5609613524ecf4f15af0f7b31abca4-Paper-Conference.pdf},
 volume = {35},
 year = {2022}
}

@inproceedings{yao2023react,
  title={{ReAct}: Synergizing Reasoning and Acting in Language Models},
  author={Yao, Shunyu and Zhao, Jeffrey and Yu, Dian and Du, Nan and Shafran, Izhak and Narasimhan, Karthik and Cao, Yuan},
  booktitle={Proceedings of the 11th International Conference on Learning Representations},
  year={2023}
}

@inproceedings{yao2024tot,
 author = {Yao, Shunyu and Yu, Dian and Zhao, Jeffrey and Shafran, Izhak and Griffiths, Tom and Cao, Yuan and Narasimhan, Karthik},
 booktitle = {Advances in Neural Information Processing Systems},
 doi = {10.52202/075280-0517},
 editor = {A. Oh and T. Naumann and A. Globerson and K. Saenko and M. Hardt and S. Levine},
 pages = {11809--11822},
 publisher = {Curran Associates, Inc.},
 title = {Tree of Thoughts: Deliberate Problem Solving with Large Language Models},
 url = {https://proceedings.neurips.cc/paper_files/paper/2023/file/271db9922b8d1f4dd7aaef84ed5ac703-Paper-Conference.pdf},
 volume = {36},
 year = {2023}
}

@article{besta2024got,
  author  = {Besta, Maciej and Blach, Nils and Kubicek, Ales and
             Gerstenberger, Robert and Podstawski, Micha{\l} and
             Gianinazzi, Lukas and Gajda, Joanna and Lehmann, Tomasz and
             Niewiadomski, Hubert and Nyczyk, Piotr and Hoefler, Torsten},
  title   = {{Graph of Thoughts}: Solving Elaborate Problems with Large
             Language Models},
  journal = {Proceedings of the AAAI Conference on Artificial Intelligence},
  volume  = {38},
  number  = {16},
  pages   = {17682--17690},
  year    = {2024},
  doi     = {10.1609/aaai.v38i16.29720},
  url     = {https://doi.org/10.1609/aaai.v38i16.29720}
}

@inproceedings{shinn2023reflexion,
 author = {Shinn, Noah and Cassano, Federico and Gopinath, Ashwin and Narasimhan, Karthik and Yao, Shunyu},
 booktitle = {Advances in Neural Information Processing Systems},
 doi = {10.52202/075280-0377},
 editor = {A. Oh and T. Naumann and A. Globerson and K. Saenko and M. Hardt and S. Levine},
 pages = {8634--8652},
 publisher = {Curran Associates, Inc.},
 title = {{Reflexion}: language agents with verbal reinforcement learning},
 url = {https://proceedings.neurips.cc/paper_files/paper/2023/file/1b44b878bb782e6954cd888628510e90-Paper-Conference.pdf},
 volume = {36},
 year = {2023}
}

@inproceedings{madaan2023selfrefine,
 author = {Madaan, Aman and Tandon, Niket and Gupta, Prakhar and Hallinan, Skyler and Gao, Luyu and Wiegreffe, Sarah and Alon, Uri and Dziri, Nouha and Prabhumoye, Shrimai and Yang, Yiming and Gupta, Shashank and Majumder, Bodhisattwa Prasad and Hermann, Katherine and Welleck, Sean and Yazdanbakhsh, Amir and Clark, Peter},
 booktitle = {Advances in Neural Information Processing Systems},
 doi = {10.52202/075280-2019},
 editor = {A. Oh and T. Naumann and A. Globerson and K. Saenko and M. Hardt and S. Levine},
 pages = {46534--46594},
 publisher = {Curran Associates, Inc.},
 title = {Self-Refine: Iterative Refinement with Self-Feedback},
 url = {https://proceedings.neurips.cc/paper_files/paper/2023/file/91edff07232fb1b55a505a9e9f6c0ff3-Paper-Conference.pdf},
 volume = {36},
 year = {2023}
}

@inproceedings{gou2024critic,
 author = {Gou, Zhibin and Shao, Zhihong and Gong, Yeyun and Shen, Yelong and Yang, Yujiu and Duan, Nan and Chen, Weizhu},
 booktitle = {International Conference on Learning Representations},
 editor = {B. Kim and Y. Yue and S. Chaudhuri and K. Fragkiadaki and M. Khan and Y. Sun},
 pages = {57734--57811},
 title = {{CRITIC}: Large Language Models Can Self-Correct with Tool-Interactive Critiquing},
 url = {https://proceedings.iclr.cc/paper_files/paper/2024/file/fef126561bbf9d4467dbb8d27334b8fe-Paper-Conference.pdf},
 volume = {2024},
 year = {2024}
}

@inproceedings{du2023improving,
  title={Improving Factuality and Reasoning in Language Models through Multiagent Debate},
  author={Du, Yilun and Li, Shuang and Torralba, Antonio and Tenenbaum, Joshua and Mordatch, Igor},
  booktitle={Proceedings of the 41st International Conference on Machine Learning},
  year={2024}
}

@inproceedings{liang2023encouraging,
    title = "Encouraging Divergent Thinking in Large Language Models through Multi-Agent Debate",
    author = "Liang, Tian  and
      He, Zhiwei  and
      Jiao, Wenxiang  and
      Wang, Xing  and
      Wang, Yan  and
      Wang, Rui  and
      Yang, Yujiu  and
      Shi, Shuming  and
      Tu, Zhaopeng",
    editor = "Al-Onaizan, Yaser  and
      Bansal, Mohit  and
      Chen, Yun-Nung",
    booktitle = "Proceedings of the 2024 Conference on Empirical Methods in Natural Language Processing",
    month = nov,
    year = "2024",
    address = "Miami, Florida, USA",
    publisher = "Association for Computational Linguistics",
    url = "https://aclanthology.org/2024.emnlp-main.992/",
    doi = "10.18653/v1/2024.emnlp-main.992",
    pages = "17889--17904"
}

@article{guo2022survey,
    title = "A Survey on Automated Fact-Checking",
    author = "Guo, Zhijiang  and
      Schlichtkrull, Michael  and
      Vlachos, Andreas",
    editor = "Roark, Brian  and
      Nenkova, Ani",
    journal = "Transactions of the Association for Computational Linguistics",
    volume = "10",
    year = "2022",
    address = "Cambridge, MA",
    publisher = "MIT Press",
    url = "https://aclanthology.org/2022.tacl-1.11/",
    doi = "10.1162/tacl_a_00454",
    pages = "178--206"
}

@inproceedings{sun2024llmqa,
  title={Harnessing multi-role capabilities of large language models for open-domain question answering},
  author={Sun, Hongda and Liu, Yuxuan and Wu, Chengwei and Yan, Haiyu and Tai, Cheng and Gao, Xin and Shang, Shuo and Yan, Rui},
  booktitle={Proceedings of the ACM Web Conference 2024},
  pages={4372--4382},
  year={2024}
}

@inproceedings{kojima2022large,
 author = {Kojima, Takeshi and Gu, Shixiang (Shane) and Reid, Machel and Matsuo, Yutaka and Iwasawa, Yusuke},
 booktitle = {Advances in Neural Information Processing Systems},
 doi = {10.52202/068431-1613},
 editor = {S. Koyejo and S. Mohamed and A. Agarwal and D. Belgrave and K. Cho and A. Oh},
 pages = {22199--22213},
 publisher = {Curran Associates, Inc.},
 title = {Large Language Models are Zero-Shot Reasoners},
 url = {https://proceedings.neurips.cc/paper_files/paper/2022/file/8bb0d291acd4acf06ef112099c16f326-Paper-Conference.pdf},
 volume = {35},
 year = {2022}
}

@inproceedings{nakov2021factcheck,
  title     = {Automated Fact-Checking for Assisting Human Fact-Checkers},
  author    = {Nakov, Preslav and Corney, David and Hasanain, Maram and Alam, Firoj and Elsayed, Tamer and Barrón-Cedeño, Alberto and Papotti, Paolo and Shaar, Shaden and Da San Martino, Giovanni},
  booktitle = {Proceedings of the Thirtieth International Joint Conference on
               Artificial Intelligence, {IJCAI-21}},
  publisher = {International Joint Conferences on Artificial Intelligence Organization},
  editor    = {Zhi-Hua Zhou},
  pages     = {4551--4558},
  year      = {2021},
  month     = {8},
  note      = {Survey Track},
  doi       = {10.24963/ijcai.2021/619},
  url       = {https://doi.org/10.24963/ijcai.2021/619},
}

@inproceedings{wang2023selfconsistency,
  title={Self-Consistency Improves Chain of Thought Reasoning in Language Models},
  author={Wang, Xuezhi and Wei, Jason and Schuurmans, Dale and Le, Quoc and Chi, Ed and Narang, Sharan and Chowdhery, Aakanksha and Zhou, Denny},
  booktitle={Proceedings of the 11th International Conference on Learning Representations},
  year={2023}
}

@inproceedings{sun2024tog,
 author = {Sun, Jiashuo and Xu, Chengjin and Tang, Lumingyuan and Wang, Saizhuo and Lin, Chen and Gong, Yeyun and Ni, Lionel and Shum, Heung-Yeung and Guo, Jian},
 booktitle = {International Conference on Learning Representations},
 editor = {B. Kim and Y. Yue and S. Chaudhuri and K. Fragkiadaki and M. Khan and Y. Sun},
 pages = {3868--3898},
 title = {Think-on-Graph: Deep and Responsible Reasoning of Large Language Model on Knowledge Graph},
 url = {https://proceedings.iclr.cc/paper_files/paper/2024/file/10a6bdcabbd5a3d36b760daa295f63c1-Paper-Conference.pdf},
 volume = {2024},
 year = {2024}
}

@inproceedings{luo2024rog,
  title={Reasoning on Graphs: Faithful and Interpretable Large Language Model Reasoning},
  author={Luo, Linhao and Li, Yuan-Fang and Haffari, Gholamreza and Pan, Shirui},
  booktitle={Proceedings of the 12th International Conference on Learning Representations},
  year={2024}
}

@inproceedings{dhuliawala2023cove,
    title = "Chain-of-Verification Reduces Hallucination in Large Language Models",
    author = "Dhuliawala, Shehzaad  and
      Komeili, Mojtaba  and
      Xu, Jing  and
      Raileanu, Roberta  and
      Li, Xian  and
      Celikyilmaz, Asli  and
      Weston, Jason",
    editor = "Ku, Lun-Wei  and
      Martins, Andre  and
      Srikumar, Vivek",
    booktitle = "Findings of the Association for Computational Linguistics: ACL 2024",
    month = aug,
    year = "2024",
    address = "Bangkok, Thailand",
    publisher = "Association for Computational Linguistics",
    url = "https://aclanthology.org/2024.findings-acl.212/",
    doi = "10.18653/v1/2024.findings-acl.212",
    pages = "3563--3578"
}

@inproceedings{baek2024kalm,
    title = "Knowledge-Augmented Language Model Verification",
    author = "Baek, Jinheon  and
      Jeong, Soyeong  and
      Kang, Minki  and
      Park, Jong  and
      Hwang, Sung",
    editor = "Bouamor, Houda  and
      Pino, Juan  and
      Bali, Kalika",
    booktitle = "Proceedings of the 2023 Conference on Empirical Methods in Natural Language Processing",
    month = dec,
    year = "2023",
    address = "Singapore",
    publisher = "Association for Computational Linguistics",
    url = "https://aclanthology.org/2023.emnlp-main.107/",
    doi = "10.18653/v1/2023.emnlp-main.107",
    pages = "1720--1736"
}

@inproceedings{hao-wu-2026-extending,
    title = "Extending First-Order Logic for Factual Reasoning over Knowledge Graphs",
    author = "Hao, Yuanzhen  and
      Wu, Desheng",
    editor = "Liakata, Maria  and
      Moreira, Viviane P.  and
      Zhang, Jiajun  and
      Jurgens, David",
    booktitle = "Proceedings of the 64th Annual Meeting of the {A}ssociation for {C}omputational {L}inguistics (Volume 1: Long Papers)",
    month = jul,
    year = "2026",
    address = "San Diego, California, United States",
    publisher = "Association for Computational Linguistics",
    url = "https://aclanthology.org/2026.acl-long.1230/",
    doi = "10.18653/v1/2026.acl-long.1230",
    pages = "26722--26738",
    ISBN = "979-8-89176-390-6"
}

\appendix

\clearpage
\appendix

% \begin{table*}[t]
% \centering
% \caption{FactKG Dataset Statistics}
% \label{tab:factkg_stats}
% \begin{tabular}{lrrrr}
% \toprule
% Type         & Written & Colloquial-Model & Colloquial-Presup & Total   \\
% \midrule
% One-hop      & 2,106   & 15,934            & 1,580              & 19,530  \\
% Conjunction  & 20,587  & 15,908            & 602                & 37,097  \\
% Existence    & 280     & 4,060             & 4,832              & 9,172   \\
% Multi-hop    & 10,239  & 16,420            & 603                & 27,262  \\
% Negation     & 1,340   & 12,466            & 1,807              & 15,613  \\
% \textbf{Total} & \textbf{34,462} & \textbf{64,788} & \textbf{9,424} & \textbf{108,674} \\
% \bottomrule
% \end{tabular}
% \end{table*}

\section{Dataset Details}
In this work, we conduct all experiments on the FactKG dataset, a challenging knowledge graph reasoning benchmark that covers diverse reasoning types and linguistic styles. The dataset contains two main linguistic categories: \textit{written} formal sentences and \textit{colloquial} informal sentences, where the colloquial set is further divided into model-generated colloquial texts and presupposition-based colloquial texts.

\begin{figure}[t]
\centering
\includegraphics[width=\linewidth]{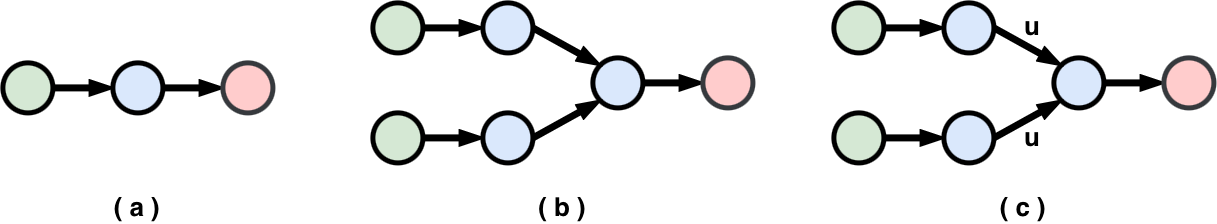}
\caption{Graph patterns for conjunction and multi-hop claims in FactKG.}
\label{fig:dataset_pattern}
\end{figure}

To comprehensively illustrate the data composition, we summarize the detailed data statistics across five typical reasoning types, including one-hop reasoning, conjunction reasoning, existence reasoning, multi-hop reasoning, and negation reasoning. Table \ref{tab:factkg_stats} presents the overall data distribution and detailed statistics. Figure \ref{fig:dataset_pattern} shows the graph patterns adopted by conjunction and multi-hop claims.

\section{Details of Baseline Methods}
\paragraph{Baselines.}
Following the evaluation protocol of \citet{kim2023factkg} and \citet{hao2025pgr}, we categorize baselines into two groups based on evidence usage. We elaborate the core design and implementation of each baseline as follows.

\begin{itemize}
\item \textbf{BERT} \citep{devlin2019bert}: A classic bidirectional Transformer model pre-trained on large-scale text corpora. We adopt the standard architecture and train it as a text classifier via full training, which takes only claim text as input without external knowledge evidence.

\item \textbf{BlueBERT} \citep{bluebert2019}: A domain-enhanced variant derived from BERT, further pre-trained on biomedical texts to strengthen semantic representation capability. Consistent with the unified setting, we train it with the full-training strategy and use pure claim text for prediction.

% \item \textbf{BlueBERT} \citep{bluebert2019}:
% A BERT-based model further pre-trained on PubMed abstracts and
% MIMIC-III clinical notes. Consistent with the unified setting, we train it with the full-training strategy and use pure claim text for prediction.

% \item \textbf{Flan-T5} \citep{chung2024flant5}: A large language model optimized via multi-task instruction tuning. We treat fact verification as a classification task and conduct full training on this model, relying merely on the original claim to make judgments.

\item \textbf{Flan-T5} \citep{chung2024flant5}: A large language model optimized via multi-task instruction tuning.
Following the original FactKG evaluation protocol, we evaluate Flan-T5
in the zero-shot setting using only the claim text.

\item \textbf{GEAR} \citep{zhou2019gear}: A graph neural network based framework specially optimized for the FactKG benchmark. It leverages knowledge graph evidence and is trained under the full-training paradigm to conduct fact verification.

\item \textbf{KG-GPT} \citep{kim2023kggpt}: A typical LLM-based knowledge graph reasoning method working under few-shot settings. It combines large language models with retrieved KG evidence to complete reasoning. We evaluate multiple model variants, including the GPT-3.5-turbo version (KG-GPT), GPT-4o version (KG-GPT$^{\dagger}$), GPT-4o-mini version (KG-GPT$^{*}$) and DeepSeek-V3 version (KG-GPT$_d$).

\item \textbf{ProgramFC} \citep{pan2023programfc}: An interpretable fact verification approach that converts natural language claims into executable logical programs. It verifies factual correctness by executing the generated programs over knowledge graph data.

\item \textbf{PGR} \citep{hao2025pgr}: A state-of-the-art program generation baseline for KG-based fact verification. We test three variants built upon different backbones, namely PGR (GPT-4o), PGR-mini (GPT-4o-mini) and PGR$_d$ (DeepSeek-V3).
\end{itemize}

% \begin{table}[t]
% \centering
% \caption{FactKG Dataset Statistics}
% \label{tab:factkg_stats}
% \resizebox{\linewidth}{!}{
% \begin{tabular}{lrrrr}
% \toprule
% Type         & Written & Colloquial-Model & Colloquial-Presup & Total   \\
% \midrule
% One-hop      & 2,106   & 15,934            & 1,580              & 19,530  \\
% Conjunction  & 20,587  & 15,908            & 602                & 37,097  \\
% Existence    & 280     & 4,060             & 4,832              & 9,172   \\
% Multi-hop    & 10,239  & 16,420            & 603                & 27,262  \\
% Negation     & 1,340   & 12,466            & 1,807              & 15,613  \\
% \textbf{Total} & \textbf{34,462} & \textbf{64,788} & \textbf{9,424} & \textbf{108,674} \\
% \bottomrule
% \end{tabular}
% }
% \end{table}

\begin{table}[t]
\centering
\resizebox{\linewidth}{!}{
\begin{tabular}{lrrrr}
\toprule
Type & Written & Colloquial-Model & Colloquial-Presup & Total \\
\midrule
One-hop      & 2,106  & 15,934 & 1,580 & 19,620 \\
Conjunction  & 20,587 & 15,908 & 602   & 37,097 \\
Existence    & 280    & 4,060  & 4,832 & 9,172  \\
Multi-hop    & 10,239 & 16,420 & 603   & 27,262 \\
Negation     & 1,340  & 12,466 & 1,807 & 15,613 \\
\textbf{Total}
& \textbf{34,552}
& \textbf{64,788}
& \textbf{9,424}
& \textbf{108,764} \\
\bottomrule
\end{tabular}
}
\caption{FactKG Dataset Statistics}
\label{tab:factkg_stats}
\end{table}

\section{Cross-Dataset Generalization}
\label{app:cross_dataset}

We further evaluate \method on Fact-FOLX-KG~\citep{hao-wu-2026-extending} to assess its transferability across datasets. Fact-FOLX-KG is a structured knowledge fact checking benchmark that requires first-order logic (FOL) reasoning across five categories.

\begin{table*}[t]
  \centering
  \footnotesize
  \setlength{\tabcolsep}{2.8pt}
  \begin{tabular}{llccccccc}
    \toprule
    \textbf{Method} & \textbf{Evidence} & \textbf{Strategy} & \textbf{Quant-Free} & \textbf{FOL-Quant} & \textbf{Entity-Comp.} & \textbf{Count-Quant} & \textbf{Count-Comp.} & \textbf{Acc.} \\
    \midrule
    GPT-4o & w/o & 0-shot & 73.25 & 59.50 & 71.77 & 60.69 & 56.15 & 62.47 \\
    DeepSeek-V3.2 & w/o & 0-shot & 66.50 & 52.47 & 64.72 & 56.05 & 52.46 & 56.70 \\
    BERT & w/o & full-training & 78.50 & 66.22 & 48.70 & 81.08 & 63.18 & 68.26 \\
    \midrule
    BERT & w/ & full-training & 92.25 & 66.37 & 52.80 & 80.43 & 61.42 & 70.12 \\
    GEAR(KG) & w/ & full-training & 91.17 & 68.45 & 55.12 & 80.81 & 66.32 & 72.50 \\
    KG-GPT$^{*}$ & w/ & few-shot & 78.00 & 47.89 & 44.22 & 48.85 & 48.96 & 51.75 \\
    KG-GPT & w/ & few-shot & 78.14 & 48.02 & 46.59 & 49.90 & 50.21 & 53.61 \\
    PGR(exec)$^{*}$ & w/ & few-shot & 93.50 & 57.47 & 50.68 & 55.15 & 59.58 & 61.19 \\
    PGR$^{*}$ & w/ & few-shot & 93.38 & 33.63 & 49.81 & 24.75 & 35.70 & 41.91 \\
    PGR & w/ & few-shot & 93.51 & 36.19 & 51.11 & 30.27 & 37.93 & 44.84 \\
    \midrule
    \textbf{DARE (Ours)} & w/ & \textbf{0-shot} & \textbf{93.62} & \textbf{77.78} & \textbf{72.26} & \textbf{83.31} & \textbf{72.41} & \textbf{79.26} \\
    \bottomrule
  \end{tabular}
  \caption{Accuracy (\%) on the Fact-FOLX-KG dataset. A superscript $*$ denotes a DeepSeek-V3.2-based baseline variant; unmarked LLM-based baseline variants use GPT-4o. For our method, we report results obtained with Qwen3-8B. The best results among the included methods are in \textbf{bold}.}
  \label{tab:factfolx}
\end{table*}

% As shown in Table~\ref{tab:factfolx}, DARE with Qwen3-8B achieves 79.26\% overall accuracy in the zero-shot setting, attaining the best performance across all five reasoning categories among the included baselines. It outperforms the strongest baseline, GEAR(KG), by 6.76 percentage points. Notably, PGR obtains 44.84\% accuracy on Fact-FOLX-KG, compared with 86.82\% on FactKG, whereas DARE achieves 79.26\% and 88.12\% on the two benchmarks, respectively. These results provide further evidence that dialectical agentic reasoning transfers effectively across datasets with distinct reasoning requirements.

\section{Prompt Templates}
\label{app:prompts}

We provide the detailed prompt templates used in DARE. 

% % 定义 Prompt 盒子样式
% \newtcolorbox{promptbox}[1]{
%   colback=gray!3!white,    % 极淡的灰色背景
%   colframe=gray!50!black,  % 深灰色边框
%   title=\textbf{#1},       % 标题加粗
%   fonttitle=\bfseries\sffamily,
%   fontupper=\small\ttfamily, % 内容使用打字机字体
%   breakable,               % 允许跨页
%   boxrule=0.8pt,
%   sharp corners=south,     % 底部直角
%   enhanced
% }

% === 颜色定义 (放在导言区或 Appendix 开头) ===
% 边框：低饱和度的深灰蓝 (Muted Slate Blue)
\definecolor{flare_frame}{RGB}{70, 110, 145} 
% 背景：极淡的冰蓝色 (Ice Blue)，保证文字清晰
\definecolor{flare_back}{RGB}{245, 248, 252} 

% === 修改后的 Prompt 盒子样式 ===
\newtcolorbox{promptbox}[1]{
  colback=flare_back,      % 替换为淡蓝背景
  colframe=flare_frame,    % 替换为深蓝边框
  coltitle=white,          % 标题文字白色
  title=\textbf{#1},       % 标题加粗
  fonttitle=\bfseries\sffamily,
  fontupper=\small\ttfamily, % 内容保持打字机字体
  breakable,               % 允许跨页
  boxrule=0.8pt,           % 线条粗细
  % sharp corners=south,     % 底部直角（顶部圆角）
  arc=2mm,
  enhanced
}

% ================= Prompt 2: Fact Checking Verification =================
\begin{promptbox}{Prompt: Dialectical Reasoning}
\#\# Instruction:\\
You are a \textbf{Fact-Checking Expert}.\\
Your task is to verify a 'Claim' using provided 'Retrieved Facts' from a Knowledge Graph and your own internal knowledge and provide two independent confidence scores (0.0 to 1.0) for a given claim.\\
\#\# GUIDELINES:\\
1.\textbf{Relevance Check}: The 'Retrieved Facts' are candidates and may be IRRELEVANT to the claim. If they are irrelevant, rely on your internal knowledge.\\
2.\textbf{Conflict Detection}: If a Retrieved Fact explicitly contradicts the claim (e.g., Claim says 'born in A', Fact says 'born in B'), the verdict is FALSE.\\
3.\textbf{Supportive Evidence}: If a Retrieved Fact confirms the claim, the verdict is TRUE.\\
4.\textbf{Negation}: Pay attention to "not", "no", etc. If the claim is negated, and the facts support the positive statement, the verdict is FALSE.\\
\#\# EVALUATION LOGIC:\\
1. \textbf{True\_Confidence}: How strongly does the Evidence and your knowledge support the claim being TRUE?\\
2. \textbf{False\_Confidence}: How strongly does the Evidence (especially entity mismatches or relation conflicts) suggest the claim is FALSE?\\
\#\# CRITICAL ADVERSARIAL CHECK:\\
- If the claim says 'Object A' but evidence shows 'Object B' for the fact, False\_Confidence should be very high.\\
\#\# FORMAT:\\
Reasoning: <Step-by-step analysis>\\
Verdict: [TRUE or FALSE]\\
- Scores: \{"True\_Confidence": float, "False\_Confidence": float\}
\end{promptbox}

\end{document}